\documentclass[runningheads]{llncs}

\usepackage{lmodern}
\usepackage[T1]{fontenc}
\usepackage{amsmath,amssymb}
\usepackage[numbers]{natbib}
\usepackage{graphicx}

\usepackage{hyperref}
\usepackage{booktabs}       
\usepackage{nicefrac}       
\usepackage{microtype}      
\usepackage{xcolor}         
\usepackage{xspace}         
\usepackage{multirow}
\usepackage{framed}         %
\usepackage{array}          
\usepackage{subcaption}
\usepackage{fontawesome5}   
\usepackage{float}          
\usepackage[most]{tcolorbox}
\usepackage{amsfonts} 
\usepackage{calc}
\usepackage{soul}
\usepackage{listings}

\definecolor{PromptBlue}{RGB}{245,245,255}
\definecolor{TitleBlue}{RGB}{0,0,153}
\definecolor{arxivRed}{RGB}{179, 28, 27}

\lstdefinestyle{promptlisting}{
  basicstyle=\ttfamily\footnotesize,
  breaklines=true,
  breakatwhitespace=false,
  columns=fullflexible,
  keepspaces=true,
  showstringspaces=false,
  frame=none
}

\newtcolorbox{promptbox}[1][]{
  enhanced,
  breakable,
  colback=PromptBlue,
  colframe=TitleBlue,
  colbacktitle=TitleBlue,
  coltitle=white,
  boxrule=1.5pt,
  rounded corners,
  fonttitle=\bfseries,
  left=0.5em, right=0.5em, top=0.5em, bottom=0.5em,
  fontupper=\small,
  fontlower=\small,
  title={#1}
}

\newtheorem{assumption}{Assumption}
\newtheorem{task}{Task}
\newtheorem{desideratum}{Desideratum}

\usepackage{color}

\newcommand{\datasetName}{\texttt{DAGverse}\xspace}
\newcommand{\SemDAG}{\textsc{SemDAG}\xspace}
\title{\datasetName: Building Document-Grounded Semantic DAGs from Scientific Papers}
\titlerunning{DAGverse}

\begin{document}
\author{Shu Wan\inst{1} \and Saketh Vishnubhatla\inst{1} \and Iskander Kushbay\inst{1} \and Tom Heffernan\inst{2} \and Aaron Belikoff\inst{2} \and Raha Moraffah\inst{2} \and Huan Liu\inst{1}}
\authorrunning{S. Wan et al.}
\institute{Arizona State University, Tempe AZ 85281, USA\\
\email{\{swan16,svishnu6,ikushbay,huanliu\}@asu.edu}
\and
Worcester Polytechnic Institute, Worcester MA 01609, USA\\
\email{\{ntheffernan,azbelikoff,rmoraffah\}@wpi.edu}}

\maketitle

\begin{abstract}
    Directed Acyclic Graphs (DAGs) are widely used to represent structured knowledge in scientific and technical domains. However, datasets for real-world DAGs remain scarce because constructing them typically requires expert interpretation of domain documents. We study Doc2SemDAG construction: recovering a preferred semantic DAG from a document together with the cited evidence and context that explain it. This problem is challenging because a document may admit multiple plausible abstractions, the intended structure is often implicit, and the supporting evidence is scattered across prose, equations, captions, and figures. To address these challenges, we leverage scientific papers containing explicit DAG figures as a natural source of supervision. In this setting, the DAG figure provides the DAG structure, while the accompanying text provides context and explanation. We introduce \datasetName, a framework for constructing document-grounded semantic DAGs from online scientific papers. Its core component, \datasetName-Pipeline, is a semi-automatic system designed to produce high-precision semantic DAG examples through figure classification, graph reconstruction, semantic grounding, and validation. As a case study, we test the framework for causal DAGs and release \datasetName-1, a dataset of 108 expert-validated semantic DAGs with graph-level, node-level, and edge-level evidence. Experiments show that \datasetName-Pipeline outperforms existing Vision-Language Models on DAG classification and annotation. \datasetName provides a foundation for document-grounded DAG benchmarks and opens new directions for studying structured reasoning grounded in real-world evidence.
\end{abstract}

\section{Introduction}

Across biostatistics, econometrics, statistics, computer science, and industry workflows, researchers and domain experts use directed acyclic graphs (DAGs) to summarize the entities, assumptions, and directional relations relevant to a specific problem~\cite{pearl2009causality,koller2009probabilistic,harenslak2021data}. In practice, these graphs are not created in isolation. They are constructed from scientific papers, technical reports, and other documents whose supporting evidence is often scattered across prose, equations, captions, and figures. This makes real-world DAG construction both labor-intensive and difficult to scale~\cite{brouillard2024landscape}, and leads to a natural question:
\begin{quotation}
Given a document, can we construct the DAG that a domain expert would draw from it?
\end{quotation}

The difficulty appears as soon as one tries to dig deeper into this question. A document may support multiple plausible DAGs. Addtionally, claims may be implicit, compressed, or scattered across sections. Consequently, the same document may lead to different resulting DAGs. One researcher may attribute a mechanism into a single node, while another may unfold it into several intermediate variables. Moreover, experts do not usually draw a graph for everything mentioned in a document. They construct a graph for a particular problem, mechanism, or figure, at a level of abstraction that is succinct for that purpose.

\begin{figure}[htbp]
  \centering
  \includegraphics[width=0.75\linewidth]{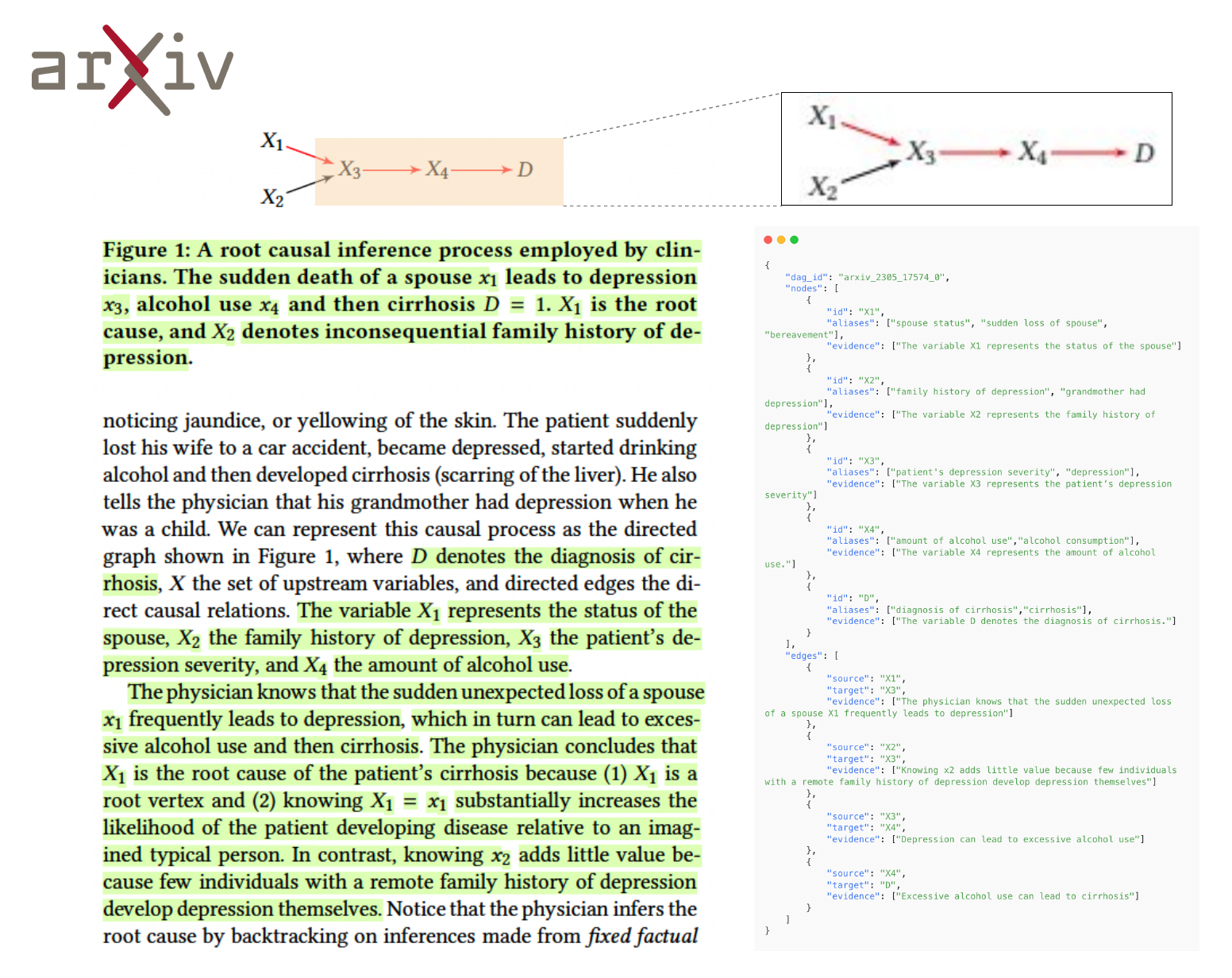}
  \caption{Constructing a semantic DAG (Appendix~\ref{app:semantic-dag}) from a scientific paper~\cite{STROBL2024104585}.}
  \label{fig:task-diagram}

\end{figure}

A bare DAG only records dependency structure. Like Wittgenstein's famous quote~\cite{wittgenstein2009philosophical}, ``the meaning of a word is its use in the language'', the meaning of a DAG is defined by its use in the document. This motivates us to extend regular DAGs by attaching evidence from the document. We call this augmented DAG a semantic DAG (\SemDAG).

This shift from bare DAGs to semantic DAGs also changes the data problem. Constructing document-grounded DAGs typically requires deep domain knowledge, specific modeling choices, and careful documentation, which makes high-quality examples difficult to obtain at scale. As a result, most existing datasets rely on synthetic or manually curated data, limiting their grounding in natural language and authentic real-world contexts. Scientific papers with DAG figures~(\ref{fig:task-diagram}) provide a favorable source of candidates for this problem. In such papers, the DAG figure provides the structure, the accompanying text often describes its structure, semantics, and application setting. With the growing availability of online scientific repositories, this creates an opportunity to build a continuously expanding dataset of document-grounded DAGs at scale.

Motivated by this opportunity, we introduce \datasetName, a framework for constructing document-grounded semantic DAGs from online scientific papers. The core component of this framework is \datasetName-Pipeline, a 5-stage semi-automated process consisting of metadata filtering, figure classification, graph reconstruction, semantic grounding, and validation. We apply this framework to causal DAGs as a concrete case study and release \datasetName-1, a benchmark featuring 108 expert-validated semantic DAGs supported by rich textual evidence.

In summary, our contributions are as follows:
\begin{enumerate}
  \item We formally define \SemDAG and the \textsc{Doc2SemDAG} task, alongside a discussion of the underlying assumptions and desiderata (Section~\ref{sec:problem-setting}).

  \item We present \datasetName-Pipeline, a modular and extensible system that uses scientific papers with DAG figures for building grounded examples for the Doc2SemDAG task (Section~\ref{sec:pipeline}).

  \item We validate the framework on causal DAGs and release \datasetName-1, a curated benchmark of 108 semantic DAGs with graph-level, node-level, and edge-level evidence (Section~\ref{sec:dataset}).

  \item We benchmark modern VLMs and our pipeline on Doc2SemDAG and show that \datasetName outperforms existing methods(Section~\ref{sec:experiments}).
\end{enumerate}

\section{Problem Setting}
\label{sec:problem-setting}

\begin{definition}[Semantic DAG]
A semantic DAG is an augmented DAG $\{G, S\}$, where $G=(V,E)$ is a DAG and $S$ contains the cited evidence and context that support how $G$ should be interpreted in the source document $D$. In our setting, $S$ may include text excerpts for nodes and edges, nodes aliases, and graph-level context such as theme and domain.
\end{definition}

In \datasetName, we represent each semantic DAG in JSON format (Appendix~\ref{app:semantic-dag}). With this target in place, we can formally define the task of contructing semantic DAGs from documents.

\begin{task}[Doc2SemDAG]
Given a document $D$ and an intended problem, recover the preferred semantic DAG $\{G, S\}$, where $S$ is grounded in $D$.
\end{task}

To make the Doc2SemDAG task well posed, we require three assumptions:

\begin{assumption}[Existence]
For the intended problem, the document admits at least one semantic DAG that captures its principal dependency structure.
\end{assumption}

\begin{assumption}[Sufficiency]
The document provides sufficient support to construct that preferred semantic DAG, including the interpretation of its nodes, edges, aliases, and graph-level context.
\end{assumption}

These assumptions should not be taken as given. Some documents do not inherently support a clean DAG abstraction. Others mention dependencies too ambiguously or omit necessary context.

Under these assumptions, a recovered semantic DAG should satisfy two desiderata.

\begin{desideratum}[Faithfulness]
A recovered semantic DAG should be faithful to the document. It should not introduce unsupported structure, and it should not omit dependencies that the document supports.
\end{desideratum}

\begin{remark}[Faithful vs.\ Factual]
Faithfulness is different from factuality. Faithfulness asks whether the semantic DAG is supported by, and self-contained within, the source document. Factuality asks whether the graph is correct with respect to external evidence or the real world. Our target is faithfulness, not factuality.
\end{remark}

\begin{desideratum}[Preferred Canonicalization]
Although multiple semantic DAGs may be compatible with a document, one canonical representation is preferred for the intended use.
\end{desideratum}

\begin{remark}[Preferred Canonicalization]
The preferred canonicalization depends on the intended problem. For example, in causal analysis, dependencies should be interpreted as actual causality~\cite{halpern2016actual}. It should avoid both over-fragmented and overly coarse representations.
\end{remark}

\subsection{Scientific Papers with DAG Figures}

The assumptions and desiderata explain why scientific papers with DAG figures are a favorable source of candidate documents. In a well-written paper, the author has already externalized a preferred structural abstraction, while the accompanying text often provides the context needed to interpret it. This does not guarantee that every paper with a DAG figure provides ground truth without validation, but it is a strong heuristic for identifying documents that are likely to support a well-defined semantic DAG. In the next section, we translate assumptions and desiderata into pipeline design principles.

\section{\datasetName-Pipeline}\label{sec:pipeline}

We introduce \datasetName-Pipeline, a semi-automatic and continuously extensible system for constructing semantic DAGs from scientific papers. The design of \datasetName-Pipeline is guided by the assumptions and desiderata in Section~\ref{sec:problem-setting}:

\begin{enumerate}
    \item \textbf{Decompose the task into specialized stages.} Doc2SemDAG combines figure understanding, graph reconstruction, and evidence grounding. Rather than ask one model to do everything, we split the problem into simpler stages that can be handled by models specialized for each step.

    \item \textbf{Prioritize precision over recall.} Errors made early in the pipeline propagate to later stages and make validation more costly. We therefore prefer to miss borderline cases rather than admit many false positives into the dataset.

    \item \textbf{Human experts as the gold standard.} Full manual construction does not scale, but fully automatic construction is still unreliable. We therefore reserve human expertise for final quality control and difficult cases, where it has the highest value.

    \item \textbf{Modular pipeline.} Each stage should be replaceable as better parsers, VLMs, or LLMs become available. This makes the pipeline suitable for continuous updates as the literature grows.
\end{enumerate}

\begin{figure}[htbp]
  \centering
  \includegraphics[width=0.7\linewidth]{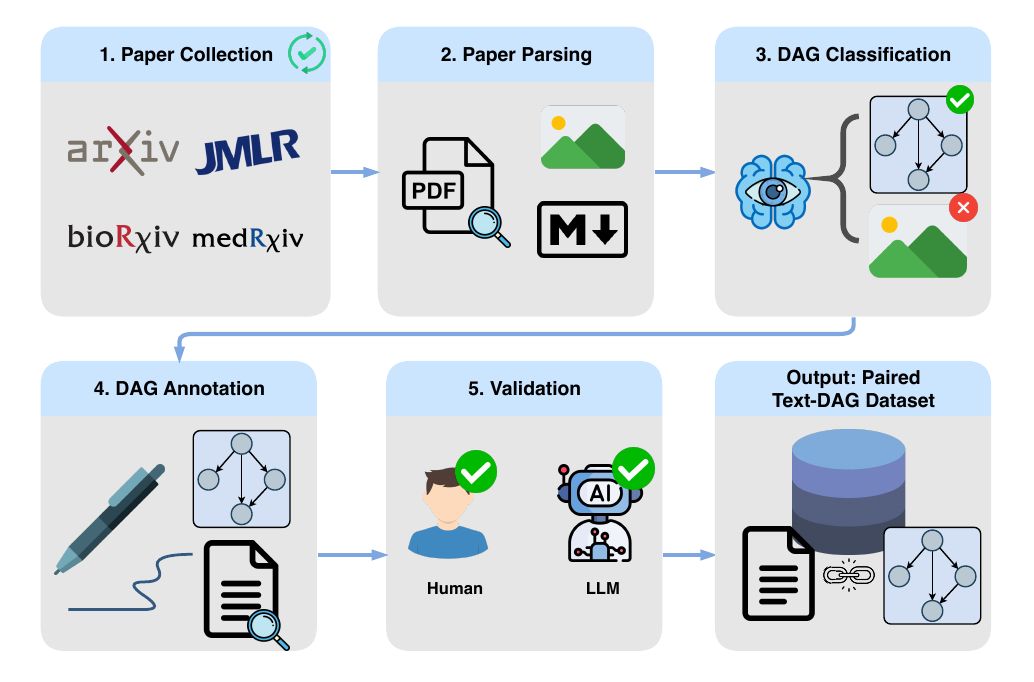}
  \caption{Overview of \datasetName-Pipeline. Paper collection can be updated routinely.}
  \label{fig:pipeline-diagram}
\end{figure}

At a high level, the pipeline (Figure~\ref{fig:pipeline-diagram}) decomposes the Doc2SemDAG task into five stages: (i) paper collection, (ii) paper parsing, (iii) DAG classification, (iv) DAG annotation \& enrichment, and (v) validation.

\datasetName-Pipeline is designed to be modular and extensible. In this paper, we use GPT-5.1~\cite{singh2025openai} for DAG classification and Gemini 2.5 Pro~\cite{comanici2025gemini} for DAG annotation.

\textbf{Paper Collection.} Public repositories host large numbers of papers, so the first step is to narrow down the search space. We utilize the metadata from these repositories to filter out irrelevant papers. We begin with simple keyword filtering, using terms such as \emph{causality}, \emph{interpretability}, and \emph{graphical models}. We then apply an LLM-based metadata filter over titles and abstracts (Appendix~\ref{app:prompts}) to retain papers that are more likely to contain applied causal case studies.

\textbf{Paper Parsing.} We parse each PDF into text blocks and extracted figures using Marker~\footnote{https://github.com/datalab-to/marker}. Text blocks include paragraphs, captions, and section headers, which we group into coherent chunks for downstream grounding. Figures are stored together with their coordinates and associated captions.

\textbf{DAG Classification.} DAG classification filters out figures that are out of scope. Given only the figure, the model decides whether it contains a single DAG. The main difficulty is to separate true DAGs from visually similar figures that also contain nodes and arrows, such as flowcharts, model diagrams, and other diagrammatic graphics. Following the prompt in Appendix~\ref{app:prompts}, the model approves only figures that depict a single connected DAG with clear directionality and rejects out-of-scope figures. This stage is tuned for high precision so that downstream annotation operates on a much cleaner candidate pool.

\begin{remark}[DAG Scope]\label{remark:dag-scope}
For simplicity, this paper focuses only on figures that contain a single, static DAG with clear directional syntax. We exclude figures that contain multiple DAGs, temporal DAGs, cyclic graphs, or mixed and ambiguous edge semantics.
\end{remark}

\textbf{DAG Annotation \& Enrichment.} After this pre-filtering step, a more capable multimodal model receives the figure together with the parsed paper text. Following the prompt in Appendix~\ref{app:prompts}, the model first re-checks whether the figure is a causal graph and assigns a coarse category. For accepted cases, it reconstructs the node set and directed edges, then produces a structured JSON annotation with graph-level context, node aliases, node and edge descriptions, and evidence snippets grounded in the parsed paper text. The model may also mark a graph as abstract when the figure lacks a concrete real-world interpretation. Candidates that fail basic structural checks, such as duplicate node IDs or invalid edges, are rejected before human review.

\datasetName-Pipeline is designed to produce semantic DAGs that support text-DAG reasoning. The enrichment schema extends with additional graph-level attributes, including the topic and domain of the DAG, and whether the DAG is technical or abstract. These fields support downstream filtering and analysis.

\section{Dataset Validation}
\label{sec:validation}

\subsection{Human Expert Validation}

\begin{figure}[htbp]
  \centering
  \begin{subfigure}[t]{0.4\linewidth}
    \centering
    \includegraphics[width=\linewidth]{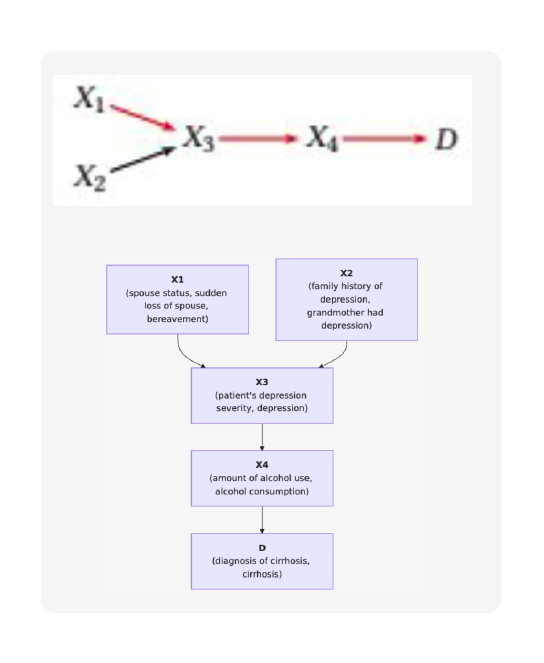}
    \caption{Reconstructed DAG view for structure checking.}
  \end{subfigure}\hfill
  \begin{subfigure}[t]{0.4\linewidth}
    \centering
    \includegraphics[width=\linewidth]{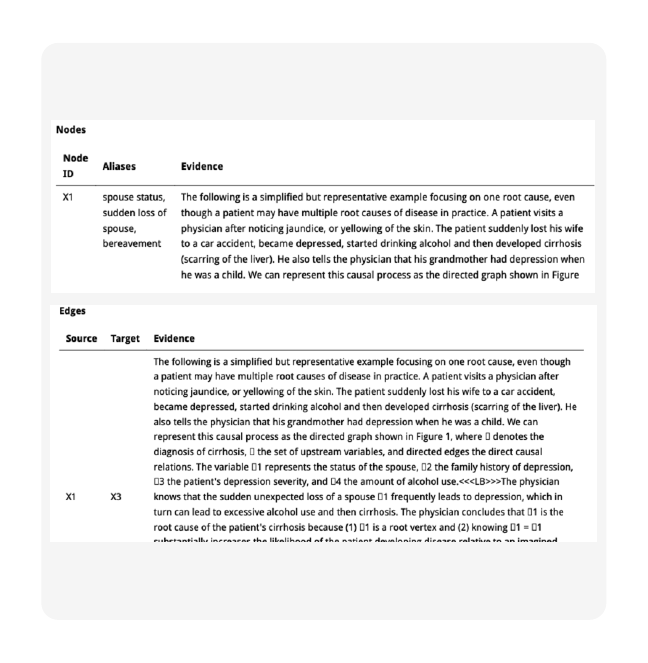}
    \caption{Evidence panel for node and edge grounding checks.}
  \end{subfigure}
  \caption{Validation interface artifacts used by experts. Left: reconstructed DAG view for structure checking. Right: evidence panel for node and edge grounding checks.}
  \label{fig:validation-gui}
\end{figure}

The automatic stages of \datasetName-Pipeline produce an initial pool of candidate semantic DAGs. Because the annotations are stored as JSON, we render each candidate into a visual review interface (Figure~\ref{fig:validation-gui}) that shows the source figure, a reconstructed graph $G$, and the corresponding evidence snippets from the parsed paper text.

Human experts then apply a two-stage decision rule. They first check whether the figure is a valid, in-scope single DAG; candidates that fail this gate are \emph{rejected} immediately. For candidates that pass, experts assess the reconstructed structure and supporting evidence together, evaluating whether node identities, edge directions, and graph-level, node-level, and edge-level evidence are accurate. We allow at most five manual edits for a candidate to be considered usable.

\subsection{LLM-as-a-Judge for Scalable Validation}
\label{sec:llm-judge}

Human validation is currently indispensable for data quality, but it does not scale well enough for a continuously updated dataset. To explore scalable alternatives, we implement a LLM-as-a-judge framework~\cite{chen2024mllm} that follows the human expert rubric. For each candidate semantic DAG, we use three independent annotator models and one judge model. The judge, drawn from a different model family, then reviews their decisions and issues a final verdict.

The judge behaves conservatively and tends to reject a subset of DAGs that humans would keep. This increases precision at the cost of recall, which is consistent with the overall design of \datasetName-Pipeline, but it also makes the system too restrictive for fully automatic validation. Inspection of the logs suggests that models are relatively strong at finding text that matches a proposed graph once the structure is given. Their main failure mode is earlier in the pipeline. They tend to be overly conservative on identifying in-scope DAG figures. This observation is consistent with the experiments in Section~\ref{sec:experiments} and supports the need for a staged pipeline rather than a single end-to-end multimodal model.

\section{Experiments}
\label{sec:experiments}

We evaluate the effectiveness and robustness of \datasetName-Pipeline on a curated benchmark of 100 paper-figure pairs drawn from causal learning papers. The benchmark contains 50 expert-validated semantic DAGs, 25 clearly non-DAG figures, and 25 DAG-like negatives that are visually similar to DAGs but fall outside our target class. The negative set includes examples such as flowcharts and model diagrams, which makes the front-end classification problem meaningfully harder than a simple graph-versus-non-graph decision.

We compare \datasetName-Pipeline against zero-shot VLM baselines that are asked to produce a semantic DAG in a single pass from the candidate figure and the corresponding paper text blocks. In contrast, \datasetName-Pipeline decomposes the task into classification, reconstruction, grounding, and validation. We report three metrics:

\textbf{DAG Classification (DC).} DC measures precision at the figure-filtering stage: among all figures predicted to contain a DAG, how many actually contain a valid single DAG. We emphasize precision because false positives propagate expensive errors into later stages of the pipeline.

\textbf{Structural Difference (SD).} For figures correctly identified as DAGs, we compare the predicted graph $\hat{G}$ against the reference graph $G$. Before computing the adjacency matrices, we align nodes between the two graphs by node identity. If a predicted node does not match any reference node, or a reference node is missing from the prediction, we treat that node as mismatched and remove all incoming and outgoing edges incident to it from the comparison. Let $A$ and $\hat{A}$ denote the resulting aligned adjacency matrices, where $A, \hat{A} \in \{0,1\}^{m \times m}$ for a graph with $m$ matched nodes. We then use one minus normalized Hamming distance so that higher is better.

\begin{equation}
  \label{eq:nhd}
  \mathrm{NHD}(G, \hat{G}) \;=\; \frac{2}{m (m - 1)} \sum_{i \ne j} \mathbf{1}_{A_{ij} \ne \hat{A}_{ij}}, \qquad
  \mathrm{SD}(G, \hat{G}) = 1 - \mathrm{NHD}(G, \hat{G})
\end{equation}

\textbf{Evidence Alignment (EA).} EA measures whether the generated node- and edge-level evidence aligns with the reference text blocks. Because the dataset is expert-validated, we can directly compare the cited block IDs. If a model fails to identify a node or edge, the corresponding evidence is counted as incorrect.

\subsection{RQ1: Pipeline Effectiveness}
\label{sec:exp-effectiveness}

To assess overall effectiveness, we compare \datasetName-Pipeline to several zero-shot VLM baselines under the metrics defined above. The table reports how many figures each method predicts as DAGs, how many of those are true single DAGs, and how many survive the full pipeline to become usable semantic DAGs.

\begin{table}[htbp]
  \centering
  \small
  \setlength{\tabcolsep}{5pt}
  \renewcommand{\arraystretch}{0.95}
  \resizebox{\linewidth}{!}{%
  \begin{tabular}{lccccccc}
    \toprule
    \textbf{Model} & \textbf{Pred.} & \textbf{True DAG} & \textbf{DC} & \textbf{SD} & \textbf{EA} & \textbf{Kept} & \textbf{E2E} \\
    \midrule
    Qwen3-VL-7B & 12 & 6 & 50\% & 52\% & 96\% & 3 & 25\% \\
    InternVL3.5-8B & 29 & 9 & 31\% & 43\% & 93\% & 7 & 24\% \\
    Qwen3-VL-30B & 21 & 12 & 57\% & 25\% & 97\% & 9 & 43\% \\
    InternVL3.5-30B & 37 & 16 & 43\% & 63\% & 98\% & 12 & 32\% \\
    GPT-5.1 & 59 & 46 & 78\% & 95\% & 98\% & 41 & 69\% \\
    Gemini2.5-Pro & 54 & 39 & 72\% & 95\% & 100\% & 39 & 72\% \\
    \hline
    DAGverse-Pipeline & 57 & 50 & 88\% & 92\% & 100\% & 50 & 88\% \\
    \bottomrule
  \end{tabular}
  }
  \caption{We compare \datasetName-Pipeline with popular open source and propretary VLMs. ``Pred.'' is the number of figures labeld as DAGs, ``True DAG'' are actual DAGs. ``DC'', ``SD'', ``EA'', and ``E2E'' are calculated with respect to each model's correct examples. Among all models, \datasetName-Pipeline has the highest retention rate, without highest number of examples kept.}
  \label{tab:pipeline-effectiveness}

\end{table}

Table~\ref{tab:pipeline-effectiveness} reveals a clear pattern. Evidence alignment is already near-saturated for modern VLMs: even smaller models score between 93\% and 98\% EA, while GPT-5.1, Gemini 2.5 Pro, and \datasetName-Pipeline reach 98\%--100\%. On this benchmark, once a model has identified the right figure and recovered a plausible graph, grounding nodes and edges back to the relevant text is comparatively easy. This suggests that evidence alignment is no longer the main bottleneck.

The harder stages are DAG classification and structure construction. Structural performance varies much more widely, from 25\% to 95\% SD, and classification precision remains far from solved. The main challenge is not distinguishing obvious non-DAG figures, which strong VLMs usually handle well, but rejecting figures that are visually similar to DAGs, such as flowcharts and model diagrams. Our logs show that these DAG-like negatives are the dominant source of errors. This is exactly where the pipeline gains its advantage: the classification stage is optimized for high-precision filtering, and the annotation stage is separated from the initial visual decision.

The strongest monolithic models are still competitive once they commit to a true DAG. GPT-5.1 keeps 41 of the 46 true DAGs that it identifies, and Gemini 2.5 Pro keeps all 39 of the true DAGs that it identifies. However, both sacrifice recall at the filtering stage by failing to retain the full set of 50 benchmark DAGs. \datasetName-Pipeline avoids this tradeoff most effectively. It achieves the best DAG classification precision (88\%), recovers all 50 usable semantic DAGs, and attains the highest end-to-end score. Taken together, these results support the necessity of a staged pipeline design. A single end-to-end VLM can often ground evidence once the input is clean, but it still struggles to robustly filter and reconstruct DAGs in realistic document collections.

\begin{table}[htbp]
  \centering
  \small
  \setlength{\tabcolsep}{5pt}
  \renewcommand{\arraystretch}{0.95}
  \begin{tabular}{lccc}
    \toprule
    \textbf{Model} & \textbf{arXiv} & \textbf{bioRxiv} \\
    \midrule
    Qwen3-VL-7B        & 5\% & 7\% \\
    InternVL3.5-8B     & 8\% & 14\% \\
    Qwen3-VL-30B       & 14\% & 43\% \\
    InternVL3.5-30B    & 21\% & 43\% \\
    GPT-5.1            & 67\% & 71\% \\
    Gemini2.5-Pro      & 73\% & 79\% \\
    \midrule
    \datasetName-Pipeline & 88\% & 93\% \\
    \bottomrule
  \end{tabular}
  \caption{End-to-end performance of \datasetName-Pipeline and baselines across different paper sources.}
  \label{tab:pipeline-robustness}

\end{table}

\subsection{RQ2: Pipeline Robustness Across Sources}
\label{sec:exp-robustness}

To study robustness, we separately evaluate papers from arXiv and bioRxiv. This tests whether the pipeline is sensitive to differences in document layout, figure style, and writing conventions across sources. In this experiment, we focus on the bottom-line end-to-end retention rate, the fraction of usable semantic DAGs recovered from each source.

Table~\ref{tab:pipeline-robustness} shows that \datasetName-Pipeline remains strong on both sources and substantially outperforms the zero-shot baselines. The performance gap is especially important because it appears before any domain-specific adaptation, suggesting that the staged design is robust to moderate changes in formatting and document style. This is necessary if \datasetName is to expand to additional venues and disciplines.

\section{\datasetName-1: A Dataset of Document-Grounded Causal DAGs}\label{sec:dataset}

\begin{figure}[htbp]
  \centering
  \begin{minipage}[c]{0.38\linewidth}
    \centering
    \scriptsize
\renewcommand{\arraystretch}{1.05}
\resizebox{\linewidth}{!}{%
  \begin{tabular}{lrrr}
    \toprule
    \textbf{Nodes} & arXiv & bioRxiv & CLadder \\
    \midrule
    Mean & 6.45 & 6.77 & 3.43 \\
    Variance & 14.53 & 20.53 & 0.25 \\
    Minimum & 2 & 3 & 3 \\
    Maximum & 25 & 19 & 4 \\
    \addlinespace[0.5em]
    \textbf{Edges} & arXiv & bioRxiv & CLadder \\
    \midrule
    Mean & 7.76 & 7.92 & 3.22 \\
    Variance & 30.19 & 29.08 & 1.23 \\
    Minimum & 1 & 4 & 2 \\
    Maximum & 32 & 19 & 5 \\
    \addlinespace[0.5em]
    \textbf{Domain} & \multicolumn{3}{r}{Count} \\
    \midrule
    Public Health & \multicolumn{3}{r}{18} \\
    Epidemiology & \multicolumn{3}{r}{17} \\
    Healthcare & \multicolumn{3}{r}{12} \\
    Education & \multicolumn{3}{r}{9} \\
    Biostatistics & \multicolumn{3}{r}{8} \\
    \bottomrule
  \end{tabular}%
}

  \end{minipage}
  \hfill
  \begin{minipage}[c]{0.6\linewidth}
    \centering
    \includegraphics[width=\linewidth]{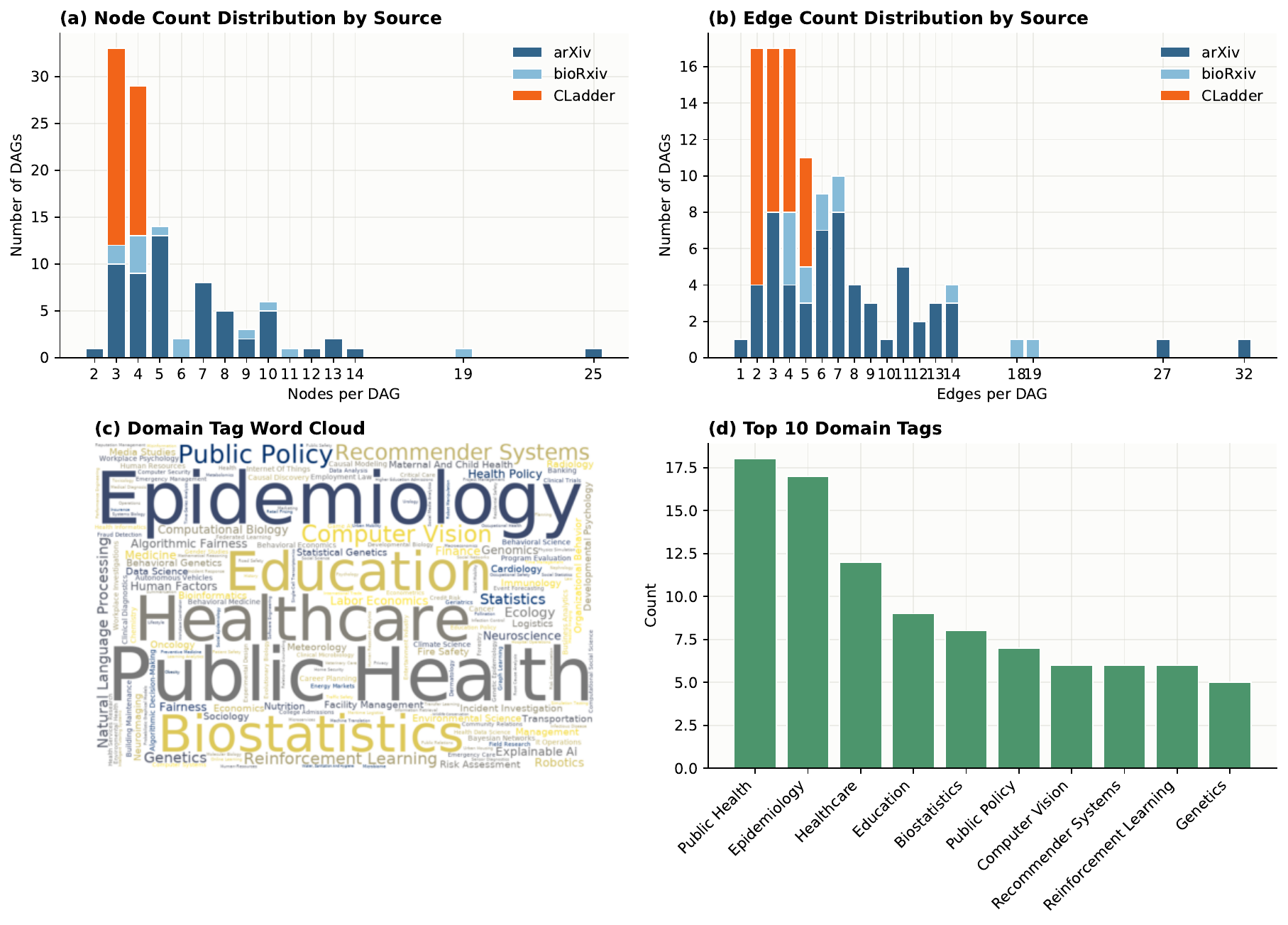}
  \end{minipage}
  \caption{\datasetName-1 data card. The left panel summarizes key statistics, including node and edge summaries and the most frequent domain tags. The right panel visualizes the node and edge distribution by source, a domain word cloud, and a top domain count plot. We omit two general tags ``Causal Inference'' and ``Machine Learning'' in domain visualizations.}
  \label{fig:dataset-diagram}

\end{figure}

We introduce \datasetName-1, the first release of the \datasetName dataset, focusing on causal DAGs. The pipeline begins with over 2.7 million arXiv papers and 400 thousand bioRxiv papers published between January 2019 and April 2025, and proceeds through a rigorous multi-stage filtering process (Table~\ref{tab:pipeline-funnel}). In addition, we incorporate synthetic DAG datasets with textual descriptions (Cladder~\cite{jin2023cladder}) to complement real-world sources. The final \datasetName-1 dataset contains 108 high-quality semantic DAGs (Figure~\ref{fig:dataset-diagram}).

Among all 108 DAGs, those sourced from arXiv are notably larger, averaging 6.51 nodes and 7.79 edges, whereas Cladder DAGs average 3.43 nodes and 3.22 edges. This distinction suggests that real-world DAGs have greater structural complexity and diversity than synthetic ones. In total, \datasetName-1 spans 190 unique domains, with the most frequent tags including public health, epidemiology, healthcare, education, and biostatistics. This concentration is largely driven by the domain proximity of bioRxiv papers, and the current domain labeling is not sufficiently granular to distinguish finer differences.

Overall, \datasetName-1 exhibits significant diversity in both domain coverage and graph structure. This distinguishes it from existing DAG datasets, which tend to be either synthetic or restricted to a narrow range of domains. The dataset is publicly available on HuggingFace~\footnote{https://huggingface.co/datasets/textual-causal-reasoning/dagverse-example}.

\subsection{Intended Use and Applications}\label{subsec:applications}

\begin{figure}[tbp]
  \centering
  \includegraphics[width=\linewidth]{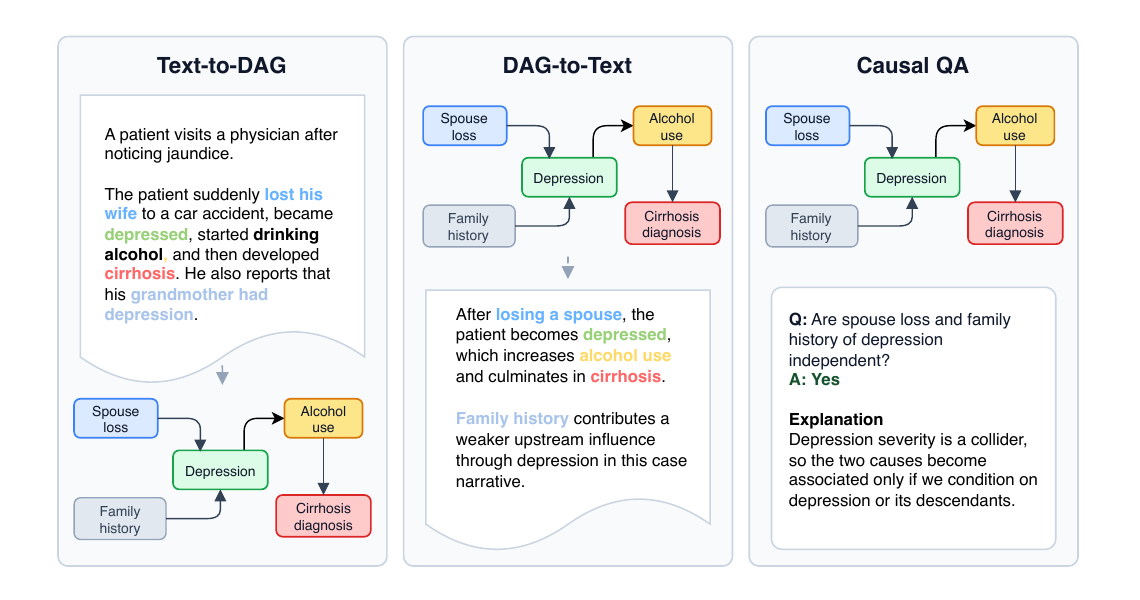}
  \caption{A single semantic DAG in \datasetName can be used in different ways. Using the cirrhosis example~\cite{STROBL2024104585}, we demonstrate 3 tasks: text-to-graph recovery from grounded evidence, graph-to-text generation of a coherent causal narrative, and causal question answering over graph structure and semantics.}
  \label{fig:applications-teaser}
\end{figure}

\datasetName-1 supports a range of tasks that require structured reasoning grounded in real-world evidence. Figure~\ref{fig:applications-teaser} shows how a semantic DAG can support different forms of text-graph reasoning.

\textbf{Text-to-DAG} generation asks a model to recover a graph from scientific prose and associated figures. The difficulty is not only relation extraction, but also choosing the right level of abstraction so that the recovered nodes and edges summarize the document-level mental model.

\textbf{DAG-to-Text} generation asks whether a model can verbalize a semantic DAG as a coherent explanation that remains globally consistent with the structure. Because each example contains graph-level, node-level, and edge-level evidence, the dataset supports evaluation beyond isolated local descriptions.

\textbf{Causal QA and reasoning}. A semantic DAG supports DAG-based reasoning because it combines graph structure with grounded meaning. For causal DAGs, the graph can act as a compact inference engine while the grounded text supplies semantic context. This enables reasoning tasks such as checking structural independence, tracing upstream causes, or testing whether a claim matches the documented graph.

\section{Related Work}

\subsection{Graph–Text Paired Datasets}

Graph–text paired datasets have been widely studied in knowledge graph verbalization and data-to-text generation. WebNLG~\cite{gardent2017webnlg} pairs small sets of DBpedia RDF triples with manually written descriptions, while WikiGraphs~\cite{wang2021wikigraphs} aligns Wikipedia passages with subgraphs from Wikidata. These resources, however, center on factual triples rather than DAGs and provide short, often stylized text that enumerates relations rather than presenting explanatory narratives. Recent work has attempted to scale up graph–text construction automatically. Mousavi et al.~\cite{mousavi2024construction} introduce LAGRANGE using a cyclic alignment framework, but find that fully automatic pairing frequently produces misaligned or incoherent text, and that even partially curated datasets, such as WebNLG, still outperform synthetic alternatives. Existing graph–text datasets highlight the difficulty of obtaining high-quality, semantically aligned graph–text pairs at scale.

\subsection{Causal DAG Datasets}

Doc2SemDAG differs from standard causal discovery, which assumes that the variables and measurements are already fixed~\cite{981e909a-69b6-346e-bd9d-1ef8e392bda3}. It also goes beyond pairwise causal relation extraction~\cite{mirza2016catena}. A wide range of causal DAG datasets support causal discovery and causal reasoning research, but most remain small, synthetic, or lack natural language grounding. Classic resources such as the \texttt{bnlearn}~\cite{scutari2010learning} offer only a few hand-crafted networks without textual descriptions. Automated extraction approaches, including Garg and Fetzer's economics-focused pipeline \cite{garg2025causal}, extract causal claims from econometric equations, producing large numbers of pairwise links. However, the resulting graphs are fragmented, domain-specific, and labeled with abstract identifiers rather than natural language concepts. Recent benchmarks shift toward evaluating LLMs' causal reasoning under controlled DAG settings. CLadder probes Pearl's hierarchy through a curated set of small graphs \cite{jin2023cladder}; Corr2Cause tests correlation–causation discrimination \cite{jin2023can}; CausalProbe examines causal signals in news text \cite{chi2024unveiling}; CausalGraph2LLM and CLEAR assess reasoning over synthetic graph scenarios \cite{sheth2024causalgraph2llm,chen2024clear}, and CausalBench aggregates multiple such evaluations \cite{zhou2024causalbench}. These datasets provide useful probes but remain limited in scale, rely on manual construction, and rarely include rich natural language context. ReCAST \cite{saklad2025can} moves closer to real-world grounding by extracting and validating causal relations from economics text, yet it is restricted to a single domain and does not yield coherent, author-provided DAGs.

\begin{table}[htbp]
  \centering
  \scriptsize
  \setlength{\tabcolsep}{3pt}
  \begin{tabular*}{\linewidth}{@{\extracolsep{\fill}}l cccc}
    \toprule
    & \multicolumn{1}{c}{DAG Statistics} & \multicolumn{3}{c}{Data Curation} \\
    \cmidrule(lr){2-2} \cmidrule(lr){3-5}
    \textbf{Dataset} & \textbf{\# Domains} & \textbf{Automatic} & \textbf{Real-world} & \textbf{Growing} \\
    \midrule
    CLadder \cite{jin2023cladder} & Synthetic & \faTimes        & \faTimes & \faTimes \\
    Corr2Cause \cite{jin2023can}  & Synthetic & \faTimes        & \faTimes & \faTimes \\
    CausalProbe \cite{chi2024unveiling} & News & \faTimes & \faCheck & \faTimes \\
    CausalGraph2LLM \cite{sheth2024causalgraph2llm} & Mostly synthetic & \faTimes        & \faTimes & \faTimes \\
    ReCAST \cite{saklad2025can}   & Econ & \faTimes & \faCheck & \faTimes \\
    CLEAR \cite{chen2024clear}    & Synthetic & \faTimes        & \faTimes & \faTimes \\
    CausalBench \cite{zhou2024causalbench} & Varied & \faTimes        & \faCheck & \faTimes \\
    \midrule
    \textbf{\datasetName (ours)} & \textbf{Multiple} & \textbf{\faCheck} & \textbf{\faCheck} & \textbf{\faCheck} \\
    \bottomrule
  \end{tabular*}
  \caption{Comparison between our dataset and existing causal DAGs datasets.}
  \label{tab:benchmark_comparison_full}

\end{table}

\subsection{LLMs for Data Curation and Graph Construction}

Our methodology leverages LLMs and VLMs as flexible components for dataset curation and graph extraction. Prior work has explored similar directions. Pan et al.~\cite{pan2023large} outline how LLMs can interface with knowledge graphs by filling gaps or translating between text and structured representations, and Long et al.~\cite{long2023can} test whether LLMs can generate causal graphs from textual descriptions, finding that models produce plausible links but often hallucinate or omit key variables. Multimodal systems extend this idea beyond text. VLMs have been applied to tasks such as chart reading \cite{pan2024flowlearn} and diagram understanding \cite{zhao2025can}, though converting full scientific figures into machine-readable graphs remains difficult. Most existing pipelines are still text-centric and require human supervision for training or verification, limiting their scalability despite the promise of LLMs and VLMs as general-purpose data-mining operators.

\section{Conclusions}
\label{sec:conclusions}

In this work, we present \datasetName, a systematic framework for constructing semantic causal DAGs from scientific documents, and release \datasetName-1, a curated dataset that bridges natural language and causal DAGs. Together, they establish a foundation for studying multimodal causal understanding and dataset construction from real-world scientific sources.

\datasetName-Pipeline is designed to prioritize precision over recall. In practice this means that ambiguous figures, noisy parses, and uncertain annotations are discarded rather than forced into a graph. This conservative design yields a trustworthy foundation for \datasetName, but it also limits coverage in the current release. Future versions of the pipeline should explore controlled ways to increase scale without losing reliability, for example by allowing multiple candidate graphs per figure, broadening the accepted DAG classes, or attaching calibrated confidence scores to intermediate decisions.

The current version of \datasetName-Pipeline depends on specific OCR, VLM, and LLM components that are not state of the art in every respect. Improvements in document layout analysis, optical character recognition, and multimodal modeling are likely to yield better figure detection, cleaner node labels, and more reliable text alignment. In the longer term, specialized end-to-end models for DAG construction from documents, which jointly reason over layout, text, and graphical structure, may outperform our modular design while using \datasetName-1 as supervision. Benchmarking multimodal LLMs and VLMs on such tasks, and defining shared splits and metrics on top of future, larger versions of \datasetName, is left for future work.

\datasetName-1 focuses on DAGs and on papers with an explicit causal or graphical modeling component. It does not yet cover other graph types such as undirected networks, dynamic graphs, or hybrid diagrams that mix causal and noncausal edges. Extending \datasetName-Pipeline to additional graph classes and broader scientific domains is an important direction for future work.

More broadly, DAGs provide a natural interface between natural text and symbolic reasoning. Recent work on multimodal document understanding and causal reasoning over scientific text~\cite{wang2024charxiv,kiciman2024causal} highlights the need for datasets and benchmarks where structured models and natural language are connected. \datasetName offers such a setting. It provides a controlled test bed for studying how models can learn to construct structured knowledge from complex documents, how they can ground that knowledge in the original text, and how they can use it for downstream reasoning tasks. We hope that \datasetName will serve as a foundation for future research on document-grounded structured reasoning and for building better tools to help researchers navigate the growing scientific literature.

\bibliographystyle{splncs04}
\bibliography{references}

\newpage
\appendix
\section{Semantic DAG}
\label{app:semantic-dag}

\begin{promptbox}[Semantic DAG Example]
\begin{lstlisting}[
  style=promptlisting,
  language={},
]
{
  "dag_id": "arxiv_2204_09274_0",
  "source": "arxiv",
  "abstract": false,
  "technical": false,
  "domain": ["healthcare", "causal inference"],
  "nodes": [
    {
      "id": "Age",
      "aliases": ["patient age", "baseline age"],
      "evidence": ["Figure 2 shows an example causal graph to study the efficiency of a medication on a disease, where the nodes represent variables and the edges represent cause-effect relations intuitively..."]
    }
  ],
  "edges": [
    {
      "source": "Age",
      "target": "Heart Rate",
      "evidence": ["Figure 2 shows an example causal graph to study the efficiency of a medication on a disease ..."]
    }
  ]
}
\end{lstlisting}
\end{promptbox}\label{fig:dag-json}

\section{Prompts}
\label{app:prompts}
\begin{promptbox}[LLM Filter Prompt]
\begin{lstlisting}[
  style=promptlisting,
  language={},
]
**Role:** You are an AI research assistant tasked with classifying academic papers based on their relevance to a specific research goal: identifying applied causal case studies rather than purely methodological papers.

**Instructions:**
Carefully follow the guideline provided below to analyze the input paper information (Category, Title, Abstract). Based *only* on the guideline and the provided paper information, determine if the paper should be classified as YES (keep) or NO (exclude).

**Guideline:**
Only include papers that demonstrate the use of causal inference techniques in a specific applied domain (e.g., economics, education, healthcare). Exclude papers that focus purely on developing new methods without an application case study.

**Input Paper Information:**
* **CATEGORY:** {category}
* **TITLE:** {paper_title}
* **ABSTRACT:** {paper_abstract}

**Output Format:**
You must respond with exactly one word: "YES" or "NO".

Answer:
\end{lstlisting}
\end{promptbox}

\begin{promptbox}[DAG Classification Prompt]
\begin{lstlisting}[
  style=promptlisting,
  language={},
]
You are an expert in causal learning. Your goal is to determine whether the image represents a single **Directed Acyclic Graph (DAG)**.

## Visual Description Guidelines

Answer `True` if the image contains a **DAG**:
- **One connected structure**: The image should display a single, unified network of nodes and arrows, not multiple disconnected groups.
- **Nodes**: Clearly separated points, circles, or labeled shapes representing entities or variables.
- **Edges**: Lines with arrows connecting nodes, indicating direction.

Answer `False` if in the image there is (are):
- No DAG: the image does not contain a single DAG
- Flowcharts: or process diagrams, decision charts, business processes, etc.
- Model diagrams: neural networks, Bayesian networks, etc.
- Undirected graphs: lines without arrows
- Graphs with cycles: arrows forming loops
- Multiple separate DAGs in one image
- Statistical diagrams: Venn diagrams, bar charts, scatter plots
- Entity-relationship diagrams: ERDs
- Circuit diagrams: electrical or logical circuits
- Ambiguous or unclear graphs: uncertain node/edge direction, unclear structure

Use `<answer>` tags for an output.

## Output Format:

```
<answer>
True/False
</answer>
```
\end{lstlisting}
\end{promptbox}

\begin{promptbox}[DAG Annotation Prompt]
\begin{lstlisting}[
  style=promptlisting,
  language={},
]
### Introduction
You are provided with a figure and the full text of a paper.
The figure is extracted from the paper and likely depicts a causal DAG (Directed Acyclic Graph).
Your task is to extract the DAG structure from the figure and enrich it using relevant information from the paper.

### Formatting
```json
You are an expert in causal learning.
Your task is to extract stories based on a provided paper and a figure extracted from the paper.
Strictly follow the instructions, using your explanation, reading comprehension, and visual understanding abilities to annotate.

## **INPUT**

- **Paper:** A causal learning paper in JSON format, with each block containing an `id` and the corresponding `text`.
- **Figure:** An image of a causal graph extracted from the paper.

**Paper:**

```
<paper>
{{paper}}
</paper>
```

## **OUTPUT FORMAT**

The output has three components:

* **label:** whether the figure meets the criteria specified in the instructions and its category
* **structure:** detailed DAG annotation
* **dags:** Semantically grounded `structure` associated with stories

Structure your output following the schema below (don't forget about (<output>, </output>) tags):
```
<output>
{
  "label":
  {
    "is_dag": true/false,
    "category": "simple" | "multiple" | "temporal" | "undetermined",
    "explanation": "..."
  },
  "structure":
  {
    "id": "{{paper_id}}",
    "nodes": [
      {
        "id": "node_id"
      }
      /* more nodes */
    ],
    "edges": [
      {
        "source": "node_id",
        "target": "node_id"
      }
      /* more edges */
    ]
  },
  "dags":
  [
    {
      "id": "{{paper_id}}_0",
      "abstract": true | false,
      "technical": true | false,         // Is understanding the DAG technical or layman-accessible?
      "domain": [], // array, default []
      "context": {
        "theme": '', // string, default ''
        "characters": [], // arrary, default []
        "setting": '', // string, default ''
        "tone": [], // array, default []
        "summary": '', // string, default ''
        "evidence": '' // array, default []
      },
      "nodes": [
        { "id": <node_id>, "aliases": [<expanded aliases, including specific or lay terms as needed>], "evidence": [paper_block_id_1,..., 'figure'] }
        /* mode nodes */
      ],
      "edges": [
        { "source": <source_id>, "target": <target_id>, "evidence": [paper_block_id_1, ..., 'figure'] }
        /* more edges */
      ]
    },
    /* one object per dag */
  ]
}
</output>
```

### Instructions

Step 1: DAG Classification.
Assess the figure to determine if it contains a causal DAG and, if so, its category. The output for this step is the label object.

1. Set `is_dag`:
  - Set to `false` if the figure contains no causal DAG (e.g., a flowchart, Venn diagram, a model diagram, or unrelated).
3. If `is_dag` is `true`, determine `category`:
   1. `multiple`: figure shows multiple separate DAGs. Use sub figure captions as cues.
   2. `temporal`: DAG describes a repeated temporal event. Cues include time indices, lags, feedback loops. Not to confuse with trivial time-dependent events.
   3. `undetermined`: contains undirected edges, bidirectional arrows, no arrows, or dashed lines. Not to confuse with unblocked paths annotations.
   4. `simple`: a single DAG, with clear nodes and edges.
4. `explanation`: provide your explanation in a few sentences.

If `is_dag` is `false`, or the `category` is in [`multiple`, `temporal`], leave the `structure` and `dags` objects empty and skip to Step 4.
If `is_dag` is `true` and `category` is in [`undetermined`, `simple`], proceed to the next step.

Step 2: Structure Extraction.
Extract essential nodes and edges from the figure

1. `id`: set to `{{paper_id}}`
2. `nodes`: create unique `id` per node, prioritize its label in the figure. If no label is available, fallback to `node_0`, etc.
3. `edges`: create directed edges with `source` and `target`, using node `id`.
4. For undetermined edges (bidirectional, no arrow, dashed line), create two separate edges (a -> b and b -> a).
5. Do not annotate minor annotations (e.g. exogenous error, logic symbol) as edges or nodes.

Step 3: Enrich the DAG `structure`.
Enrich the `structure` with information from paper. It's possible more than one story can be associated with a single DAG.
- The output should be about the DAG and its narrative only; ignore unrelated parts of the paper.
- Fill the `dags` array with one object for each distinct DAG identified.
- `id`: set to `{{paper_id}}_N`, where N starts from 0.
- `abstract`:
   -  Set to `true` if the DAG is a methodological, abstract, or theoretical diagram where nodes are trivial (e.g., confounder, treatment) and do not represent real-world entities. Otherwise, set to `false`.
- `technical`:
   -  `true` if the DAG requires domain expertise to interpret, otherwise false.
   -  `false` if the DAG is layman-accessible, even though some nodes use technical terms, such as drug names.
- `domain`: List all relevant research or application domains for the DAG as an array (e.g., ['economics', 'healthcare']). Leave empty if indeterminate.
- `context`:
   - `theme`: The phenomenon, mechanism, or story of the DAG.
   - `characters`: list of agents or actors in the DAG story, empty array if with no clear actors or the DAG is technical.
   - `setting`: Describe the relevant scenario, time, or place implied by the DAG.
   - `tone`: A list of compatible and varied tones fit the DAG's story (e.g. neutral, somber, happy, hopeful, etc.).
   - `summary`: Write a brief, readable narrative of the DAG's story, referencing node concepts but not all edges. Start with 'The story ...', do not use the word 'DAG'.
   - `evidence`: Array of paper_block_ids from paper and/or the string 'figure'
- For each `node` within the `dags` object:
   - `id`: Reference the id created in Step 2.
   - `aliases`: Extend aliases as needed for self-containment, from generic to concrete. IF the DAG is technical, and node represents a specific domain concept, do not extend, unless the paper support it.
- For each `edge` within the `dags` object:
   - `source` and `target`: Reference the ids created in Step 2.
- Evidence: For all evidence fields (nodes, edges, context), list only the supporting paper block IDs and/or the string 'figure'. Do not include textual content.

Step 4: Validate
Narrative Integrity
- Each dag object should be understandable as a self-contained narrative.
- Evidence arrays must contain valid paper_block_ids or the string 'figure'.
DAG integrity
- All node IDs must be unique.
- The number of nodes in structure and dags must match the essential nodes in the figure.
- All edge sources and targets must reference valid node IDs.

### Task

#### Paper Input
Paper: A paper in JSON format, with each block containing an `id` and the corresponding `text`.
use `id` to reference in the evidence.
Example:
```json
  {
    "id": 48,
    "text": "...",
    "rules": [
      "..."
    ]
  }
```

Paper:

{{paper}}
\end{lstlisting}
\end{promptbox}

\section{Pipeline Funnel}
\label{app:pipeline-funnel}

\begin{table}[htbp]
  \centering
  \small
  \setlength{\tabcolsep}{6pt}
  \renewcommand{\arraystretch}{1.1}

  \begin{tabular}{lrr}
    \toprule
    Number of Instances & \textbf{arXiv} &\textbf{bioRxiv} \\
    \midrule
    Metadata & 2{,}720{,}631 (100\%) & 401{,}231 (100\%) \\
    Metadata (processed) & 13{,}441 (0.49\%) & 1{,}187 (0.30\%) \\
    Papers (Downloaded) & 8{,}410 (0.31\%) & 693 (0.17\%) \\
    Papers (Candidates) & 260 (0.0096\%) & 688 (0.17\%) \\
    Figures (Pre-DAG classification) & 2{,}233 (0.08\%) & 9{,}261 (2.31\%) \\
    Figures (Post-DAG classification) & 1{,}552 (0.06\%) & 2{,}169 (0.54\%) \\
    Semantic DAGs (Pre-validation) & 66 (0.0024\%) & 14 (0.0035\%) \\
    Semantic DAGs (Validated) & 58 (0.0021\%) & 13 (0.0032\%) \\
    \bottomrule
  \end{tabular}

  \caption{End-to-end pipeline funnel from raw metadata to validated outputs. For each source, the percentage in parentheses is the retention rate at that step relative to the total number of papers in that source's metadata.}
  \label{tab:pipeline-funnel}

\end{table}




\section{Ethical Considerations}
\label{app:ethical-legal}
\datasetName reflects the papers it is built from. The DAGs and associated text are faithful to the source papers in the sense that they capture what authors wrote and drew, but they are not guaranteed to be factually correct about the real world. Some graphs describe hypothetical scenarios, stylized examples, or modeling assumptions that may not hold outside their original context. Users should therefore treat \datasetName as a resource for studying how models interact with scientific arguments and representations, not as an authority on domain facts.

\datasetName-1 is constructed from open-access repositories such as arXiv, and bioRxiv, and we respect the terms of use of each source. We release only derived annotations, metadata, and limited figure crops, leaving original PDFs hosted by the publishers, and intend the dataset for research use only. Because the corpus is skewed toward computer science, preprints, and English-language papers, users should be cautious when generalizing findings to other domains and languages.

\section{Disclosure of Gen-AI Usage}
\label{app:gen-ai-disclosure}
Generative AI tools were used during the development of this paper to assist with coding implementation and writing. All research ideas, methodology design, and manuscript writing are original work by the authors.

\end{document}